\documentclass{article}

\usepackage{PRIMEarxiv}

\usepackage[ruled]{algorithm2e} 

\usepackage{amsmath}
\usepackage{amssymb}
\usepackage{mathtools}
\usepackage{amsthm}
\usepackage{hyperref}
\usepackage{url}
\usepackage{media9}
\usepackage{siunitx}
\usepackage{pifont}
\usepackage{lipsum}
\usepackage{amsmath}
\usepackage{stfloats}
\usepackage{enumitem}
\usepackage{tcolorbox}
\usepackage{ragged2e}
\usepackage{wrapfig}
\usepackage{multirow}
\usepackage{adjustbox}
\usepackage{booktabs}
\usepackage{graphicx}
\usepackage{makecell}
\usepackage{tcolorbox}
\usepackage{ragged2e}
\usepackage{tabularx}
\usepackage{booktabs}
\usepackage{float}

\tcbuselibrary{listings,breakable,skins}

\newcommand\mypara[1]{\vspace{1pt}\noindent\textbf{#1.}}

\title{Social Structure Matters in 3D Human-Human Interaction Generation}

\author{
  Zhongju Wang \\
  University of New South Wales \\
  \texttt{zywang9691@gmail.com} \\
  \And
   Beier Wang \\
   University of New South Wales \\
  \texttt{beier.wang@unsw.edu.au} \\
  \And
   Yatao Bian \\
   National University of Singapore\\
  \texttt{ybian@nus.edu.sg} \\
  \And
   Pichao WANG \\
   NVIDIA \\
  \texttt{pichaowang@gmail.com} \\
  \And
  Zhi Wang \\
  Nanjing University \\
  \texttt{zhiwang@nju.edu.cn} \\
   \And
   Daoyi Dong \\
   University of Technology Sydney \\
  \texttt{daoyidong@gmail.com} \\
  \And
   Hongdong Li \\
   Australian National University \\
  \texttt{hongdong.li@anu.edu.au} \\
   \And
   Huadong Mo\thanks{Corresponding Author} \\
   University of New South Wales \\
  \texttt{huadong.mo@unsw.edu.au} \\
  \And
   Zhenhong Sun \\
   Australian National University \\
  \texttt{zhenhongsun1992@outlook.com} \\
}

\begin{document}
\maketitle

\vspace{-20pt}
\begin{center}
{\large \textbf{Code:} \href{https://github.com/EngineeringAI-LAB/SocialStructureHHI}{https://github.com/EngineeringAI-LAB/SocialStructureHHI}}
\end{center}
\vspace{20pt}

\begin{abstract}
Although text-to-motion generation has achieved strong progress in synthesizing realistic single-person motions from language, extending it to text-driven 3D human-human interaction (HHI) remains non-trivial, as HHI requires modeling the underlying \textbf{social structure} that governs phase progression, actor roles, and inter-actor coordination.
In this paper, we formulate HHI generation as a social structure modeling and grounding problem: the model must first infer how an interaction unfolds and how the two actors coordinate their roles, and then realize this structure as continuous, physically plausible, and partner-aware 3D motion. 
To study how such structure should be modeled, we first examine the capability boundary of large language models (LLMs) for HHI generation. 
Our analysis shows that LLMs can \textit{think} by recovering phase decompositions and partner-aware roles, but cannot directly \textit{move}, as they fail to generate dynamic, physically plausible, and interaction-aware motion. 
This motivates our planner-executor paradigm, \textbf{Think with LLM, Move with Motion Skill}. 
The LLM planner converts implicit interaction semantics into motion-aligned social supervision by decomposing interactions into phases, assigning partner-aware actor roles, and aligning them with motion sequence. 
The motion executor then grounds the planned social structure into coordinated two-person motion by adapting a pretrained solo motion model with LoRA, previous-phase self-conditioning, and ego-relative partner conditioning. 
Together, our Solo-to-Social framework bridges social organization and motion realization, producing 3D HHI with improved phase consistency, role alignment, and partner-aware coordination.
\end{abstract}

\section{Introduction}
\label{sec:intro}
Text-to-motion generation has made rapid progress in synthesizing realistic single-person 3D motion from natural language~\cite{sui2026survey, sahili2025text, fan20253d}. 
Recent large-scale motion generators further show that pretrained motion backbones can learn strong atomic motion priors from broad single-person motion data~\cite{hymotion2025, azadi2023make}. 
As generative models move toward embodied AI~\cite{liang2025large}, multi-agent collaboration~\cite{wu2025generative}, and social robotics~\cite{matheus2025long}, the research focus is naturally shifting from individual motion synthesis to text-driven 3D human-human interaction (HHI) generation. 
In this setting, a model is expected to generate two coordinated human motions from a global interaction description, such as one person approaching another person, hugging them, and then releasing the hug. 
Compared with single-person motion generation, HHI generation requires not only realistic individual motion, but also coherent coordination between two actors over time.

\begin{figure}[t]
    \centering
    \includegraphics[width=1\linewidth]{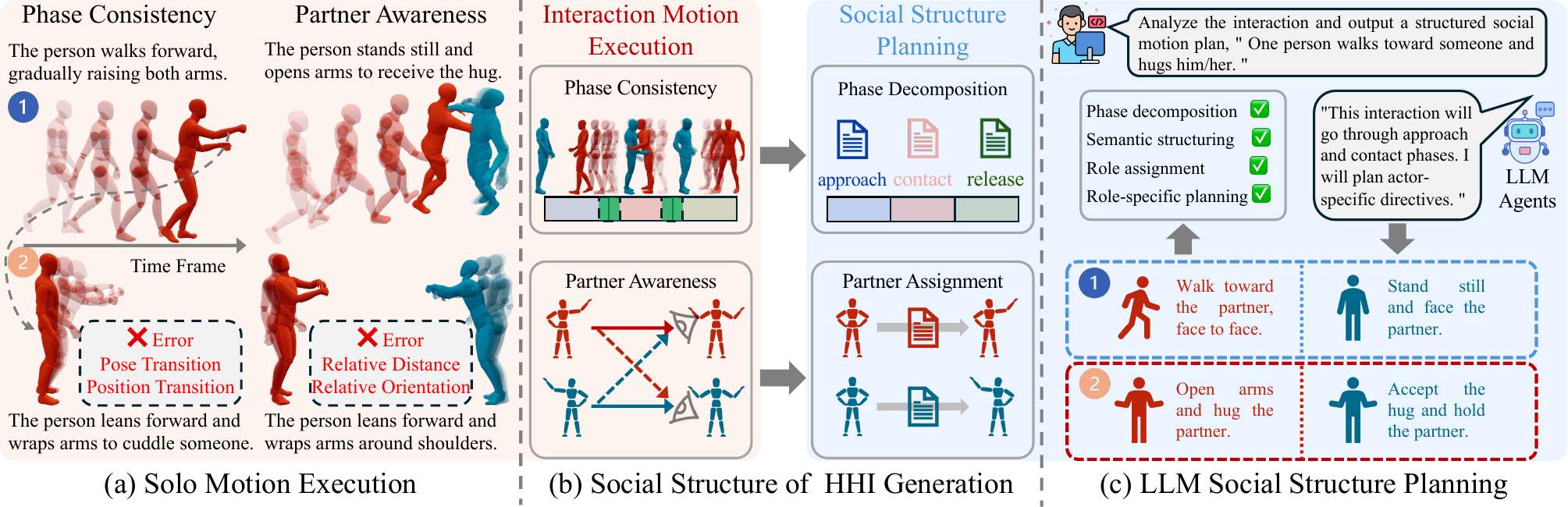}
    \caption{Social structure of text-driven HHI generation. 
    (a) Solo motion execution provides strong intra-personal motion priors but lacks interaction-level coordination.
    (b) HHI requires social structure along two dimensions: phase-level temporal organization and partner-aware coordination.
    (c) LLMs offer social planning abilities to make such structure explicit for interaction motion execution.}
    \label{fig:motivation}
\end{figure}

Existing HHI generation methods have made promising progress by jointly modeling two-person motion, composing individual motion priors, or introducing interaction-aware generation mechanisms~\cite{Ruiz_Ponce_2024_CVPR, ruiz2025mixermdm, shafir2024human, liang2024intergen, javed2025intermask, wang2025timotion}. However, treating HHI as a direct extension from one actor to two overlooks a key property of HHI: it is not merely a spatial combination of two plausible individual motions, but is organized by an underlying \textbf{social structure}, as shown in Fig.~\ref{fig:motivation}. We define social structure as the latent interaction organization that governs how an interaction unfolds over time and how two actors coordinate their roles with respect to each other. It contains two essential dimensions: \textit{phase progression}, which describes the temporal stages of an interaction, such as approach, contact, release, or in-place coordination; and \textit{partner-aware coordination}, which captures the asymmetric but coupled responsibilities of the two actors in each phase, such as initiator and receiver, attacker and defender, or giver and taker. Without such structure, generated motions may appear plausible for each actor but fail as an interaction: the actors may approach at inconsistent times, miss the contact moment, face the wrong direction, or execute incompatible roles.

This observation suggests that the core difficulty of text-driven HHI generation lies not only in motion realism, but also in recovering the social structure that guides two actors into coherent interactions. We therefore formulate HHI generation as a \textbf{social structure modeling and grounding} problem: the model must infer the interaction organization implied by a global language description, including phase decomposition and partner-aware role assignment, and ground it into continuous, physically plausible, and partner-aware 3D motion. In this view, phase consistency and partner awareness are not merely evaluation properties, but direct outcomes of correctly grounded social structure. Large Language Models (LLMs) are natural tools for making such structure explicit, given their strong language understanding~\cite{team2026qwen3}, commonsense reasoning~\cite{merlo2025human}, and structured planning~\cite{zou2025llm} abilities. Yet whether they can serve as complete HHI generators remains unclear. We therefore conduct a diagnostic study by representing Skinned Multi-Person Linear (SMPL)~\cite{SMPL:2015} parametric human motion in a token-like form and asking an LLM to generate interaction sequences from text. The results reveal a clear separation between \textit{thinking} and \textit{moving}: the LLM can infer plausible phase decompositions and partner-aware roles, but fails to reliably produce continuous, dynamic, and physically plausible 3D interaction motion. Thus, LLMs are suitable as social structure planners, but not as direct motion executors.

Based on this insight, we propose a planner-executor paradigm for social-structure-centered HHI generation: \textbf{Think with LLM, Move with Motion Skill}. The LLM acts as a social structure planner, while a pretrained motion model serves as an executable motion skill. Instead of using the LLM as a direct motion generator, we use it to convert implicit interaction semantics into explicit social structure supervision. Given a global interaction prompt and paired motion sequence, we decompose the interaction into phase-level units, assign partner-aware roles to both actors, and align each phase with its motion segment. This converts coarse HHI text-motion pairs into fine-grained, motion-aligned social annotations, making social structure a trainable bridge between language intent and motion execution.
To ground this structure into continuous two-person motion, we introduce a \textit{Solo-to-Social} (S2S) motion execution framework, which adapts a pretrained solo motion backbone into an interaction motion executor rather than training from scratch. S2S preserves the atomic motion prior learned from single-person data while adding the social coordination ability required by HHI. Phase-wise self motion conditioning uses previous-phase motion prefixes to encourage smooth transitions and long-range consistency, while partner-aware motion conditioning injects the partner's latest motion into the actor's ego-centric frame to model relative position, orientation, and interaction geometry. With parameter-efficient Low-Rank Adaptation (LoRA), these mechanisms turn a solo motion model into a socially aware executor for coherent two-person motion.
Experiments on standard HHI benchmarks show improved text-motion alignment, phase consistency, and partner-aware coordination over existing baselines, with qualitative results showing clearer phase progression, more role-consistent behaviors, and more plausible inter-person geometry. Overall, our framework treats social structure as the central abstraction of HHI generation, planned by LLM reasoning and grounded through motion skill adaptation.

The main contributions of this paper are summarized as follows:
\begin{itemize}[leftmargin=*, noitemsep, nolistsep]
    \item[$\bullet$] We identify \textbf{social structure} as a central abstraction for text-driven 3D HHI generation, and formulate HHI generation as a social structure modeling and grounding problem rather than a direct extension of solo motion generation.
    \item[$\bullet$] We analyze the capability boundary of LLMs for HHI generation, showing that LLMs are effective at social structure planning but inadequate for direct continuous 3D motion execution.
    \item[$\bullet$] We propose an LLM-based social structure planning strategy that reorganizes HHI datasets into fine-grained motion-aligned annotations with explicit phase structure and partner-aware roles.
    \item[$\bullet$] We introduce a Solo-to-Social motion execution framework that adapts a pretrained solo motion backbone into a socially aware interaction motion skill through phase-wise self conditioning, ego-relative partner conditioning, and parameter-efficient LoRA adaptation.
\end{itemize}

\section{Related Work}
\label{sec:related_work}
\mypara{Text-to-Motion Generation}
Text-to-motion generation has been widely studied for synthesizing single-person 3D motion from natural-language descriptions~\cite{zhu2023human, khani2025motion, chen2025language}. 
Early methods adopt GANs~\cite{amballa2025ls, wang2020learning}, VAEs~\cite{guo2020action2motion, zhong2023attt2m, zhang2023generating}, or autoregressive Transformers~\cite{xu2023actformer, jiang2024motiongpt, wang2024motiongpt, zhu2025motiongpt3}, while recent diffusion models and large-scale pretrained motion backbones achieve stronger realism, diversity, and text-motion alignment~\cite{tevet2022human, karunratanakul2023guided, gong2023tm2d, zhang2024motiondiffuse, guo2022tm2t, petrovich2023tmr, azadi2023make, hymotion2025}. 
These models provide powerful atomic motion priors for individual human dynamics. 
However, they are primarily designed to align one actor's motion with text, and therefore do not explicitly model the phase progression and partner-aware coordination required by human-human interaction. In contrast, text-driven HHI generation requires not only realistic individual motion, but also explicit coordination between two actors. We therefore preserve the atomic motion capacity of pretrained solo models while introducing social-structure conditioning that guides them toward coordinated two-person interaction.

\mypara{Human-Human Interaction Generation} Human-human interaction generation extends motion synthesis from individual behavior to coordinated two-person motion~\cite{stergiou2018understanding, fan20253d, sui2026survey}. 
Existing works include reaction generation, which predicts one actor's response to another actor's motion~\cite{xu2024regennet, siyao2022bailando, yao2023dance, siyao2024duolando, li2024interdance, tan2025think, cen2025ready_to_react}, and text-driven HHI generation, which produces two-person motion from language descriptions~\cite{ghosh2024remos, chopin2023interaction, xu2024inter, ruiz2025interact2ar, park2025unified, Goel2022intermix, wu2025intermamba, geng2026disentangled, ruiz2025mixermdm}. 
Representative methods such as 
ComMDM~\cite{shafir2024human}, in2IN~\cite{Ruiz_Ponce_2024_CVPR}, InterGen~\cite{liang2024intergen}, InterMask~\cite{javed2025intermask}, and TIMotion~\cite{wang2025timotion} improve two-person motion synthesis through joint modeling, motion-prior composition, or interaction-aware generation mechanisms. 
Despite these advances, most methods still treat HHI mainly as a two-person motion generation problem, without explicitly exposing the latent social structure that organizes an interaction into temporal phases and asymmetric but coupled actor roles. In contrast, we formulate HHI generation as social structure modeling and executing, where phase progression and partner-aware coordination serve as explicit intermediate representations.

\mypara{LLM-based Planning for Motion and Interaction}
LLMs exhibit strong language understanding, commonsense reasoning, and structured planning abilities~\cite{team2026qwen3, merlo2025human, zou2025llm}, making them useful for decomposing high-level human intentions into organized plans. For HHI, such reasoning can reveal how an event unfolds and how actors should coordinate their roles. 
However, continuous 3D motion execution requires kinematic precision, temporal smoothness, and physically plausible inter-person geometry, which remain difficult for LLMs to generate directly. We therefore use LLMs not as motion generators, but as social structure planners: the LLM converts implicit interaction semantics into phase-level and role-level supervision, while a pretrained motion model grounds this structure into coordinated two-person motion. 
This planner-executor view connects the semantic reasoning strength of LLMs with the motion prior of specialized motion generators.

\section{Methodology}
\label{sec:methodology}
\begin{figure}[t]
    \centering
    \includegraphics[width=1\linewidth]{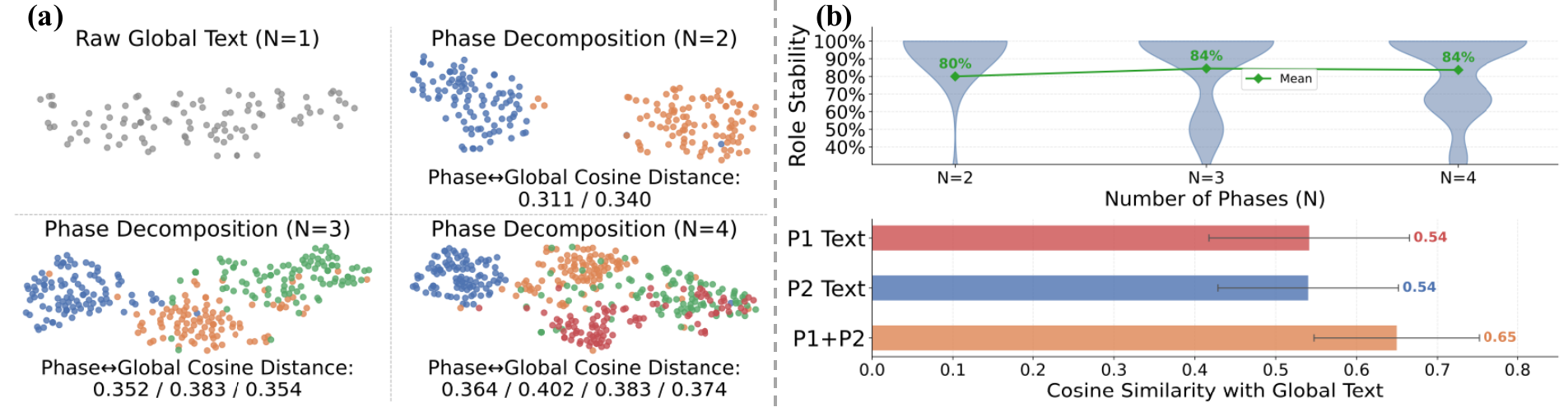}
    \caption{LLM (Qwen3.5) capability analysis in modeling social structures. (a) t-SNE of phase decomposition. The LLM can recover latent phase progression from global text. (b) Role assignment. Decoupled individual semantics are cross-phase consistent and aligned with the global intent. These results indicate that LLMs can model social structure of HHI well at the semantic level.}
    \label{fig:llm_analysis}
\end{figure}

\subsection{LLM Capability Boundary: Social Structure Planning vs. Motion Execution}
\label{subsec:method_llm_analysis}
Given a natural-language description $y$, text-driven HHI generation aims to synthesize a two-person motion sequence
\begin{equation}
p\!\left(\mathbf{X}^{1:T}\mid y\right), \qquad 
\mathbf{X}^{1:T}=\big\{\mathbf{x}^{(1)}_t,\mathbf{x}^{(2)}_t\big\}_{t=1}^{T},
\end{equation}
where $\mathbf{x}^{(i)}_t\in\mathbb{R}^{D}$ denotes the SMPL-based motion representation of actor $i$ at frame $t$. 

We define the social structure as a latent intermediate variable $S$ that captures sufficient interaction organization, the text-driven HHI generation problem can be reformulated as 
\begin{equation}
p(\mathbf{X}^{1:T}\mid y)
=
\sum_S p(\mathbf{X}^{1:T},S\mid y)
=
\sum_S p(\mathbf{X}^{1:T}\mid S,y)\,p(S\mid y)
\approx
\sum_S p_\theta(\mathbf{X}^{1:T}\mid S)\,p_\phi(S\mid y)
,
\end{equation}
where $p_{\phi}(S\mid y)$ models the social structure planning while 
$p_{\theta}\!\left(\mathbf{X}^{1:T}\mid S\right)$ models the motion execution. However, since whether LLMs can directly serve for both  $p_{\phi}(S\mid y)$ and $p_{\theta}\!\left(\mathbf{X}^{1:T}\mid S\right)$ still remains unclear, we first conduct a diagnostic study along these two dimensions. To make such analysis reproducible and experimentally controlled, we employ Qwen3.5~\cite{team2026qwen3}, an open-weight LLM with standard local deployment support, which allows the analysis to be conducted under fixed model parameters and inference settings, without introducing additional variability from service-side API updates or access policies.

\begin{wrapfigure}{r}{0.4\textwidth} 
    \centering
    \vspace{-10pt}
    \includegraphics[width=0.4\textwidth]{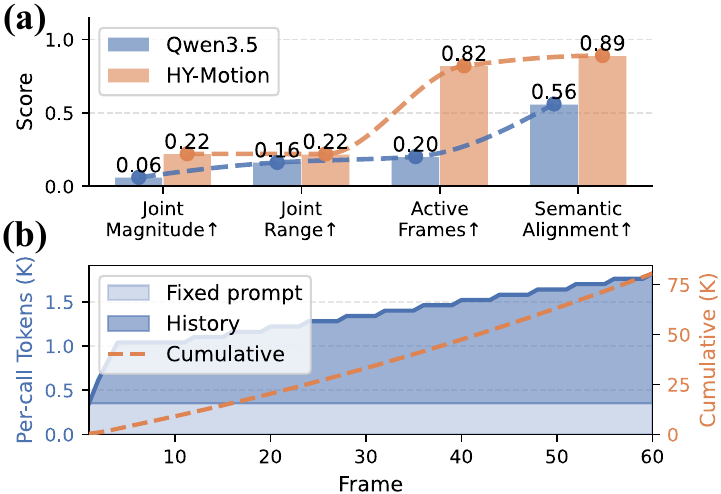}
    \caption{(a) Atomic motion generation capacity. (b) Reasoning efficiency}
    \vspace{-1pt}
    \label{fig:llm_motion_execute}
\end{wrapfigure}
\mypara{Social Structure Planning}
We first test whether LLMs can model $p_{\phi}(S\mid y)$ to capture latent social structure from global interaction text. As shown in Fig.~\ref{fig:llm_analysis}(a), Qwen3.5 decomposes a raw global prompt into increasingly separable semantic phases, indicating its ability to impose temporal structure on a coarse interaction description.
Fig.~\ref{fig:llm_analysis}(b) further shows that Qwen3.5 can infer asymmetric actor roles with high cross-phase consistency, and joint role reasoning better preserves the global interaction semantics than assigning each actor independently. These results suggest that LLMs are effective social planners for HHI, especially in resolving phase progression and partner-conditioned actor roles.

\begin{figure}[t]
    \centering
    \includegraphics[width=1\linewidth]{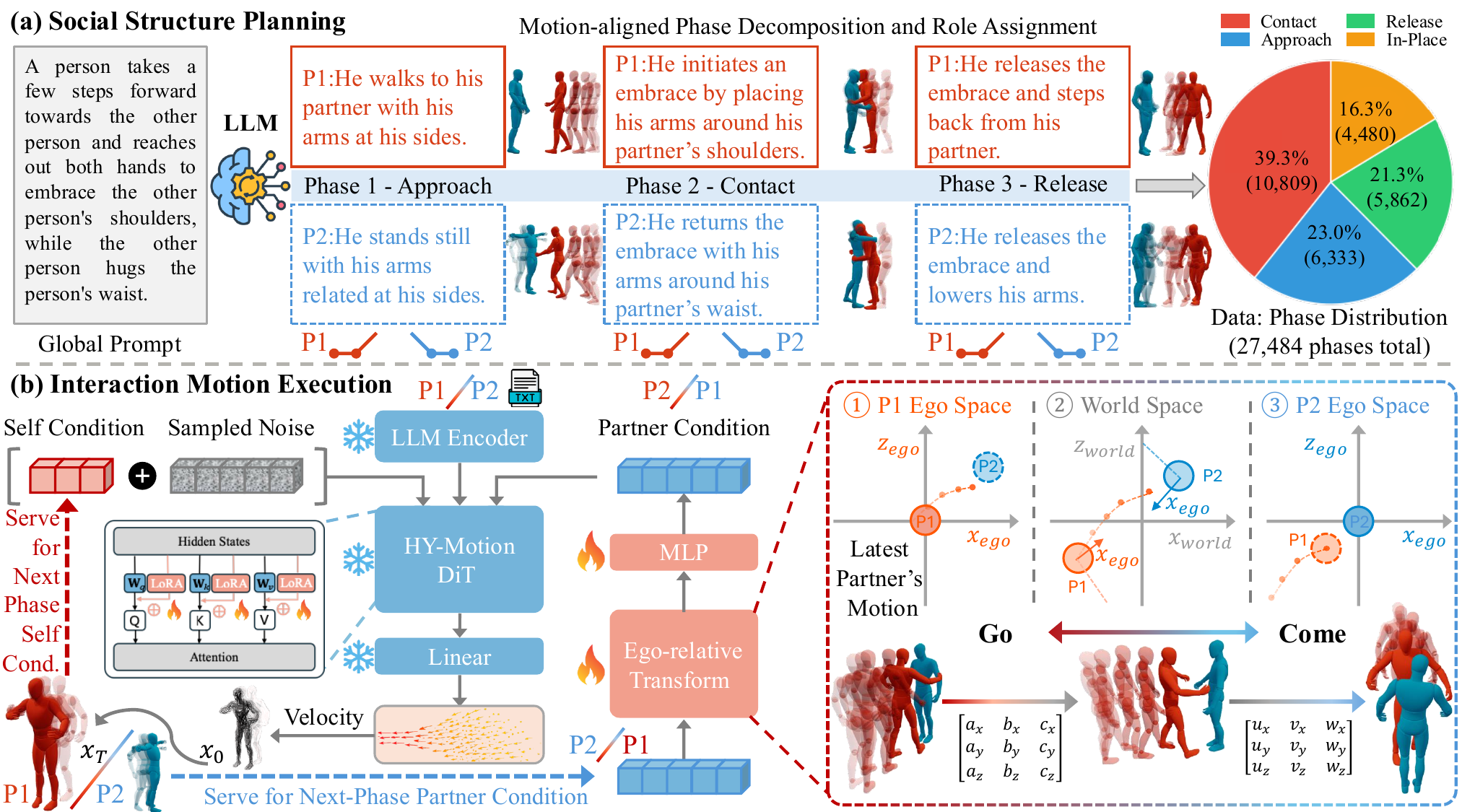}
    \caption{Overview of our proposed planner-executor paradigm for social-structure-centered HHI generation.
    (a) The LLM serves as a social structure planner which recovers plausible phase decompositions and partner-aware role assignments from global prompt.
    (b) The motion skill is built on a solo motion backbone equipped with self and partner conditioning for motion execution.
    }
    \label{fig:method_overall}
\end{figure}

\mypara{Direct Motion Execution}
We then examine whether LLMs can model $p_{\theta}\!\left(\mathbf{X}^{1:T}\mid S\right)$ for motion execution. As shown in Fig.~\ref{fig:llm_motion_execute}(a), Qwen3.5 remains weaker than a dedicated motion model even when motion is represented in the structured SMPL space. It produces limited active motion, with low joint magnitude, narrow motion range, and fewer active frames. Although the generated motion may retain partial semantic alignment with the prompt, it lacks the continuity and dynamics required for physically plausible HHI, with visualizations provided in \textbf{Appendix~\ref{app:llm_atomic_motion_execution}}. Moreover, Fig.~\ref{fig:llm_motion_execute}(b) further shows that autoregressive LLM-based motion execution incurs substantial cumulative context growth, making long-horizon motion generation increasingly inefficient.
These results reveal a clear capability boundary: LLMs can expose the social structure of HHI, but cannot reliably execute it as continuous interaction motion. This motivates a planner-executor paradigm for social-structure-centered HHI generation: \textbf{Think with LLM, Move with Motion Skill}, as shown in Fig.~\ref{fig:method_overall}. In this paradigm, the LLM serves as a social structure planner, while a pretrained solo motion model serves as an executable motion skill. In the following sections, we introduce details of these two components.

\subsection{Motion-aligned Social Structure Planning}
\label{subsec:method_social_planning}

Social structure $S$ provides the temporal and role-level organization needed for coherent HHI generation, but such supervision is not explicitly available in existing HHI datasets. 
Most datasets pair a global interaction description with a full two-person motion sequence, leaving phase progression and partner-aware actor roles implicit. To obtain the social structure $S$ based on real motion data, we reconstruct existing HHI data into motion-aligned social supervision through our LLM-based social structure planning. To prevent the LLM from producing language-plausible but motion-inconsistent plans, we extract motion facts from real motion data and use it as constraints during planning. 

Given a paired HHI motion sequence \(\mathbf{X}^{1:T}\), we first detect phases that the interaction goes through
\begin{equation}
\mathcal{C}=f_{\mathrm{p}}(\mathbf{X}^{1:T}), 
\qquad 
\mathcal{C}=\{c_k\}_{k=1}^{K}, \quad c_k\in\mathcal{V}_{\text{phase}},
\end{equation}
where $c_k$ is the $k$-th intermediate phase that the interaction goes through, the phase vocabulary is $ \mathcal{V}_{\text{phase}}=\{\texttt{approach},\ \texttt{contact},\ \texttt{release},\ \texttt{in\mbox{-}place}\}$, and the mapping \(f_{\mathrm{p}}(\cdot)\) is implemented by a forward state machine over motion-derived signals. In this way, the full sequence is decomposed into an explicit phase progression, exposing progress of the interaction, as shown in Fig.~\ref{fig:method_overall}(a).

For each phase \(c_k\), we extract a structured motion fact representation from the phase-wise motion sequence, represented as
\begin{equation}
\mathcal{M}_k=\{m_k^{(1)}, m_k^{(2)}\}, \quad
m_k^{(i)}=
\big(
\delta_k,\,
u_k,\,
q_k,\,
\mathbf{a}_k^{(i)},\,
\mathbf{l}_k^{(i)}
\big), \quad i\in\{1, 2\},
\end{equation}
where \(\delta_k\) denotes inter-person distance evolution, \(u_k\) denotes the motion initiator, \(q_k\) captures contact semantics, \(\mathbf{a}_k^{(i)}\) summarizes the motion direction and facing state of actor \(i\), and \(\mathbf{l}_k^{(i)}\) describes limb-level cues such as reaching, arm lifting, bending, and foot activity. All descriptors are computed directly from 3D motion geometry, so that \(m_k^{(i)}\) remains explicitly grounded in observable interaction motion behaviors.

Conditioned on the global interaction text \(y\), phase label \(c_k\), and motion facts \(\mathcal{M}_k\), the LLM planner reasons actor-specific semantic descriptions,
\begin{equation}
\mathcal{Y}_k \sim p_{\phi}\!\left(\mathcal{Y}_k \mid y,\, c_k,\, \mathcal{M}_k\right), \qquad \mathcal{Y}_k=(y_k^{(1)},y_k^{(2)}),
\end{equation}
where $(y_k^{(1)}, y_k^{(2)})$ represents actor-specific prompts.

Finally, aggregating all phases yields a motion-aligned social structure
\begin{equation}
S=\{(c_k, y_k^{(1)}, y_k^{(2)})\}_{k=1}^{K}.
\end{equation}
Although the final \(S\) contains only phase labels and actor-specific semantic descriptions, it is constructed under phase-wise motion fact constraints and is therefore aligned with the underlying motion sequence. 
Thus, \(S\) is not merely a plausible textual plan, but a motion-aligned intermediate representation that connects phase progression and actor-specific role semantics to the observed interaction motion, making it directly usable for downstream motion execution.

During training, paired motion facts are available and used to constrain the LLM planner. 
The resulting social structures are therefore guided by real motion evidence rather than free-form language plausibility alone, providing temporally aligned supervision for training the downstream executor. 
This process also serves as a fine-grained HHI annotation pipeline, converting coarse global text-motion pairs into phase-level and role-aware training annotations. 
At inference time, ground-truth motion facts are unavailable, so the planner operates purely from the global interaction prompt. 
This allows the LLM to exploit its semantic generalization ability to infer plausible social structure directly from language, without being tied to a specific observed motion instance, thereby preserving both high-level interaction organization and generation diversity.

\subsection{Interaction Motion Execution}
\label{subsec:method_motion_executor}

Given the structured social plan recovered by the LLM planner, the remaining challenge is to ground it into continuous and coordinated 3D human-human motion. 
We therefore introduce a Solo-to-Social (S2S) motion execution framework, which adapts a pretrained solo motion backbone into a social interaction executor, as shown in Fig.~\ref{fig:method_overall}(b). 
The core idea of S2S is to preserve the atomic motion prior learned from large-scale motion corpus, while injecting the missing mechanisms needed to execute social structure: previous-phase self-conditioning grounds phase progression into continuous motion, ego-relative partner conditioning grounds partner-aware coordination into inter-person geometry, and LoRA adaptation transfers the solo motion prior toward social execution.

\mypara{Grounding Phase Progression via Self Conditioning}
The social plan decomposes an interaction into temporally ordered phases. 
However, executing each phase independently would produce discontinuities at phase boundaries, breaking the intended phase progression. 
To ground phase progression into a continuous motion trajectory, we introduce phase-wise self motion conditioning. 
Specifically, for the $k$-th phase, its self condition is taken from the last \(e\) frames of the phase $k-1$:
\begin{equation}
\mathbf{x}_{k}^{(i)}[1:e]
=
\mathbf{x}_{k-1}^{(i)}[T_{k-1}-e+1:T_{k-1}],
\label{eq:anchor_prefix}
\end{equation}
where \(T_{k-1}\) denotes the duration of phase \(k-1\). 
The anchor region serves as a self-conditioning signal and is taken from the last \(e\) frames of the previous phase, and the model predicts the remaining frames of the current phase. 
In this way, each phase is executed with explicit access to the actor's self recent motion history, encouraging smooth phase transitions.

\mypara{Grounding Partner-aware Coordination via Partner Conditioning}
Beyond phase continuity, HHI requires each actor to move with awareness of the partner's behavior. 
A solo motion backbone operating on each actor independently cannot directly capture relative displacement, facing direction, contact timing, or interaction geometry. 
To ground partner-aware coordination, we condition the generation of actor \(i\) on the partner's motion context:
\begin{equation}
p_{\theta}\!\left(\mathbf{x}_{k}^{(i)} \mid y_k^{(i)},\, c_k,\, \mathbf{p}_{k}^{(i)}\right),
\qquad
\mathbf{p}_{k}^{(i)}
=
R_{j\rightarrow i}\!\left(\mathbf{x}_{k}^{(j)}\right),
\qquad j\neq i,
\label{eq:partner_conditioned_phase_generation}
\end{equation}
where \(y_k^{(i)}\) denotes the actor-specific role description in phase \(k\), \(c_k\) is the phase label, and \(\mathbf{p}_{k}^{(i)}\) is the partner-aware conditioning signal. \(R_{j\rightarrow i}(\cdot)\) transforms actor \(j\)'s motion into actor \(i\)'s ego-centric coordinate frame. 
This ego-relative representation makes the partner's position, orientation, and motion dynamics directly observable to the current actor.
During inference, S2S is instantiated as two coupled actor-wise executors, one for each actor. 
The two motions are rolled out synchronously in a phase-by-phase manner according to the planned phase sequence. 
For phase \(k\), each executor uses its own previous-phase motion as the self condition and the latest generated motion state of the other actor as the partner condition after ego-relative transformation. 

\mypara{Transferring Solo Motion Priors to Social Interaction}
We employ HY-Motion 1.0 ~\cite{hymotion2025}, a state-of-the-art pretrained solo motion generation model, as the core backbone of our motion executor. Instead of training the HHI generator from scratch, we adapt this solo backbone for social interaction execution using parameter-efficient LoRA modules. 
The overall fine-tuning objective is
\begin{equation}
\mathcal{L}_{\mathrm{total}}
=
\mathcal{L}_{\mathrm{FM}}
+
w_{\mathrm{smooth}}\mathcal{L}_{\mathrm{smooth}}
+
w_{\mathrm{dist}}\mathcal{L}_{\mathrm{rel\_dist}}
+
w_{\mathrm{ori}}\mathcal{L}_{\mathrm{rel\_ori}},
\end{equation}
where \(\mathcal{L}_{\mathrm{FM}}\) is the original flow-matching objective of HY-Motion 1.0 and preserves intra-person motion quality. 
\(\mathcal{L}_{\mathrm{smooth}}\) penalizes abrupt frame-to-frame changes in the generated region, encouraging temporally smooth phase execution. 
\(\mathcal{L}_{\mathrm{rel\_dist}}\) and \(\mathcal{L}_{\mathrm{rel\_ori}}\) constrain the generated actor to match the target partner-relative distance and orientation, respectively, thereby grounding partner-aware coordination in inter-person geometry. More details are provided in \textbf{Appendix~\ref{app:modeling_derivation}}.

\begin{table}[t]
    \centering
    \caption{Quantitative evaluation results.
    Abbreviations: Plan + HYM1 = Our social structure planning with HY-Motion 1.0 model, without proper grounding. 
    MM and MM Dist. denote Multimodality and Multimodal Distance, respectively. FID compares ground-truth and generated motion distributions in a normalized motion feature space, while MM Dist. averages text-motion embedding distances in a shared normalized feature space. User study details are provided in~\textbf{Appendix \ref{app:user_study}}.}
    \label{tab:Quantitative_results}
    \renewcommand{\arraystretch}{1.08}
    \setlength{\tabcolsep}{2.1pt}
    \scriptsize
    \resizebox{\linewidth}{!}{
    \begin{tabular}{@{}l ccc ccc cccc ccc@{}}
        \toprule
        \multirow{2}{*}{\textbf{Method}}
        & \multicolumn{3}{c}{\textbf{R Prec.} (\%) $\uparrow$}
        & \multicolumn{3}{c}{\textbf{FID} $\downarrow$}
        & \multicolumn{4}{c}{\textbf{User Study} $\downarrow$}
        & \multicolumn{3}{c}{\textbf{Text-Motion Statistics}} \\
        \cmidrule(lr){2-4}
        \cmidrule(lr){5-7}
        \cmidrule(lr){8-11}
        \cmidrule(lr){12-14}
        & Top 1 & Top 2 & Top 3
        & P1 & P2 & Avg.
        & Global & Partner & Phase & Avg.
        & MM $\uparrow$ & MM Dist. $\downarrow$ & Diversity $\uparrow$ \\
        \midrule

        Ground Truth 
        & 26.86 & 45.84 & 59.68
        & -- & -- & --
        & -- & -- & -- & --
        & -- & 0.94 & 8.09 \\ 

        ComMDM~\cite{shafir2024human}
        & 15.21 & 28.83 & 43.49
        & 0.90 & 0.91 & 0.90
        & 5.11 & 5.03 & 5.04 & 5.06
        & 0.56 & 0.99 & 5.38 \\

        in2IN~\cite{Ruiz_Ponce_2024_CVPR}
        & 17.34 & 30.05 & 43.64
        & 0.78 & 0.78 & 0.78
        & 5.15 & 5.20 & 5.10 & 5.15
        & 1.12 & \textbf{0.94} & 7.20 \\

        InterGen~\cite{liang2024intergen}
        & 17.34 & 31.23 & 43.87
        & 0.76 & 0.76 & 0.76
        & \underline{3.50} & 3.58 & 3.51 & 3.53
        & 1.28 & \textbf{0.94} & 7.28 \\

        InterMask~\cite{javed2025intermask}
        & 17.23 & 30.96 & 43.82
        & 0.78 & 0.74 & 0.76
        & \textbf{3.02} & \underline{3.10} & \underline{3.38} & \underline{3.17}
        & 1.41 & \underline{0.96} & 7.26 \\

        TIMotion~\cite{wang2025timotion}
        & 17.07 & 31.90 & 45.52
        & 0.71 & \underline{0.70} & \underline{0.70}
        & 4.45 & 4.53 & 4.51 & 4.49
        & 1.01 & \textbf{0.94} & \textbf{7.53} \\

        \midrule

        Plan + HYM1
        & \underline{20.25} & \underline{34.81} & \underline{48.84}
        & \underline{0.69} & \underline{0.70} & \underline{0.70}
        & 3.75 & 3.78 & 3.89 & 3.81
        & \underline{1.50} & \underline{0.96} & 7.24 \\

        Ours
        & \textbf{24.67} & \textbf{42.34} & \textbf{55.80}
        & \textbf{0.65} & \textbf{0.67} & \textbf{0.66}
        & \textbf{3.02} & \textbf{2.78} & \textbf{2.57} & \textbf{2.79}
        & \textbf{1.67} & \textbf{0.94} & \underline{7.31} \\

        \bottomrule
    \end{tabular}
    }
\end{table}

\section{Experiments}
\subsection{Implementation Details}
\label{subsec:implement_details}
\mypara{Baselines and Metrics}
We compare our S2S framework with five conventional two-person motion generation methods, including (1) ComMDM~\cite{shafir2024human}, (2) in2IN~\cite{Ruiz_Ponce_2024_CVPR}, (3) InterGen~\cite{liang2024intergen}, (4) InterMask~\cite{javed2025intermask}, and (5) TIMotion~\cite{wang2025timotion}, and one social structure baseline that adapts our social structure planning to raw HY-Motion 1.0 model (Plan + HYM1)~\cite{hymotion2025}. Conventional two-person motion generation baselines are conditioned on the global prompt, while HYM1 is conditioned on phase prompts derived by our social structure planning. All methods are evaluated across four dimensions: text-motion retrieval (R-Precision), motion realism (FID), human preference (user study), and text-motion statistics (Multimodality, MM Dist, and Diversity). More details regarding the metrics and user study are provided in \textbf{Appendix~\ref{app:metric_details} and \ref{app:user_study}}.

\mypara{Datasets} We use InterHuman~\cite{liang2024intergen} and InterX~\cite{xu2024inter}, two widely adopted SMPL-based HHI datasets. We reorganize them with our motion-aligned social structure planning. The resulting fine-grained HHI training set contains 27,484 phase segments, each paired with a phase label, partner-aware role descriptions, and the corresponding motion sequence. Evaluation is conducted on 914 sequences that consist of 454 InterHuman cases and 460 InterX cases.

\mypara{Setup} We initialize the motion executor with the open-source HY-Motion-1.0-Lite weights and finetune it with LoRA ($r=16$ and $\alpha=32$), which introduces only approximately 8M trainable parameters (1.7\% of the backbone parameter size). We train S2S for 100 epochs with a batch size of 64, with AdamW (learning rate $1\times10^{-4}$ and weight decay $1\times10^{-3}$) on an H100 GPU cluster. The number of flow-matching steps is set to 50, and the predicted frame length is 300 with 30 fps. The loss contribution ratio is set to 7:1:1:1 for $\mathcal{L}_{\mathrm{FM}}$, $\mathcal{L}_{\mathrm{smooth}}$, $\mathcal{L}_{\mathrm{rel\_dist}}$, and $\mathcal{L}_{\mathrm{rel\_ori}}$, respectively.

\begin{figure}
    \centering
    \includegraphics[width=1\linewidth]{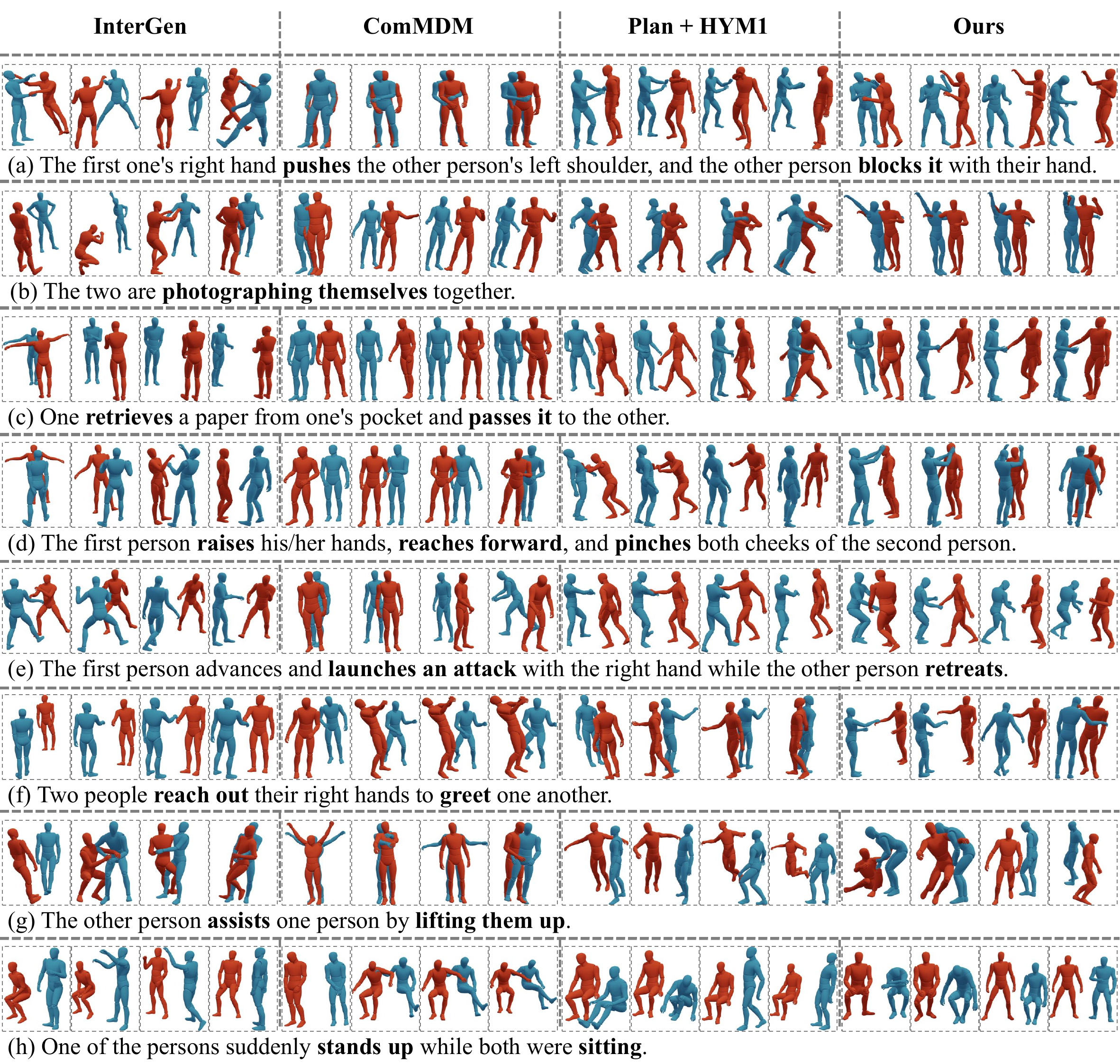}
    \caption{Qualitative evaluation results. We compare InterGen, ComMDM, a baseline using only our structure planning with the raw HY-Motion 1.0 model (Plan + HYM1), and Ours (Our structure planning + Our S2S framework). 
    The social structure planning results used to condition HYM1 and our S2S are provided in \textbf{Appendix~\ref{app:social_planning}}.
    }
    \label{fig:main_quali_res}
\end{figure}

\subsection{Main Results}

\mypara{Quantitative Evaluation}
Table~\ref{tab:Quantitative_results} reports quantitative evaluation results of all methods. 
Overall, the results show that our method improves HHI generation from two complementary perspectives: \textit{phase progression} and \textit{partner-aware coordination}. 
From the perspective of \textit{phase progression}, our method achieves the best R-Precision across Top-1/2/3, indicating stronger alignment with global interaction semantics and better preservation of the intended temporal event order. 
This advantage is also reflected in the user study, where our method obtains the best Phase ranking of 2.57, suggesting clearer phase-level temporal progression. 
From the perspective of \textit{partner-aware coordination}, our method obtains the lowest average FID, showing that grounding social structure improves interaction organization while maintaining motion quality. 
The user study further confirms this advantage, with our method achieving the best Partner ranking of 2.78, indicating stronger mutual responsiveness and more coherent partner-aware interaction quality. 
Competitive Multimodality and Diversity further show that these gains are achieved while preserving diverse motion generation. 
Together, these results demonstrate that planning and grounding social structure improves both temporal phase organization and inter-person coordination in text-driven HHI generation.

\mypara{Qualitative Evaluation} Fig.~\ref{fig:main_quali_res} provides rendered interaction motions across representative methods. 
Existing methods often generate plausible individual poses but fail to organize them into coherent interaction phases, as seen in pushing (a), greeting (f), and lifting (g), where baselines miss key stages such as approach, contact, or release, leading to premature contact, static interaction, or incomplete action progression. 
They also exhibit weak partner-aware coordination: the receiver may not block the push, the partner may not respond to a handshake, or the lifted person may not coordinate with the supporter. 
In contrast, our method produces clearer phase transitions and better preserves asymmetric but coupled roles, such as attacker--defender (e), giver--receiver (c), and lifter--assisted person (g). 
Compared with Plan + HYM1, our results further show that social planning alone is insufficient, and that the Solo-to-Social executor is necessary to ground planned phases and roles into plausible two-person geometry and coordinated motion.

\begin{table}[t]
    \centering
    \caption{Ablation study results. R-Prec.Top 1 denotes R-Precision Top 1. FID is Fr\'echet Inception Distance. PTS, RDE, ROE, and IPR represent Phase Transition Smoothness, inter-person Relative Distance Error, inter-person Relative Orientation Error, and Inter-Penetration Rate, respectively. }
    \label{tab:ablation_results}
    \renewcommand{\arraystretch}{1.08}
    \setlength{\tabcolsep}{2.1pt}
    \scriptsize
    \resizebox{\linewidth}{!}{
    \begin{tabular}{lccccccc}
    \hline
    Ablation & R Prec.Top 1$\uparrow$ & FID$\downarrow$ & PTS$\downarrow$ & RDE$\downarrow$ & ROE$\downarrow$ & IPR$\downarrow$ & Diversity$\uparrow$ \\ \hline
    w/o social structure planning & 15.25\% & \multicolumn{1}{l}{0.78} & 0.61 & 0.41 & 0.39 & 0.21 & 6.61 \\
    w/o motion facts & 23.61\% & 0.74 & 0.62 & 0.34 & 0.34 & 0.12 & \textbf{7.85} \\
    w/o self condition & 20.65\% & \multicolumn{1}{l}{0.70} & 0.71 & 0.40 & 0.29 & 0.12 & 7.18 \\
    w/o partner condition & 21.03\% & 0.69 & 0.62 & 0.61 & 0.50 & 0.14 & 7.26 \\
    Ours & \textbf{24.67\%} & \textbf{0.66} & \textbf{0.58} & \textbf{0.31} & \textbf{0.27} & \textbf{0.10} & 7.31 \\ \hline
    \end{tabular}
}
\end{table}

\subsection{Ablation Study}
We conduct ablation studies to examine how each component contributes to social structure modeling and grounding, with details provided in \textbf{Appendix~\ref{app:ablation_visualization}}.
As shown in Table~\ref{tab:ablation_results}, removing social structure planning causes the largest R-Precision drop, from 24.67\% to 15.25\%, showing that global text alone lacks fine-grained interaction supervision. 
Removing motion facts worsens FID and PTS, indicating that language-only planning may produce motion-inconsistent phase structures. 
For grounding, removing self conditioning mainly hurts phase progression, increasing PTS from 0.58 to 0.71, while removing partner conditioning damages inter-person coordination, increasing RDE from 0.31 to 0.61 and ROE from 0.27 to 0.50. 
These results show that both motion-aligned social structure planning and proper motion grounding are necessary for HHI generation.

\section{Conclusion}

In this paper, we formulate text-driven 3D human-human interaction generation as a social structure modeling and grounding problem. 
Instead of treating HHI as a direct two-person extension of solo motion generation, we identify phase progression and partner-aware coordination as the key organizing principles behind coherent interactions. We identify a clear capability boundary of LLMs: they can recover high-level social structure, but cannot reliably execute continuous and physically plausible interaction motion. 
We then propose a planner-executor paradigm, \textbf{Think with LLM, Move with Motion Skill}, where an LLM planner reconstructs motion-aligned social supervision from existing HHI data, and a Solo-to-Social executor adapts a pretrained solo motion model to ground the planned structure into coordinated two-person motion. 
Experiments on standard HHI benchmarks demonstrate that our method improves text-motion alignment, phase consistency, and partner-aware coordination, validating the importance of both social structure planning and motion-level grounding. 
We hope this work encourages future research to model social organization explicitly when building generative systems for interactive human motion and socially intelligent embodied agents.


\bibliographystyle{unsrt}  
\bibliography{my_bib}

\newpage
\appendix

\section{Metric Details}
\label{app:metric_details}
\begin{itemize}[leftmargin=*, noitemsep, nolistsep]
\item[$\bullet$]
\textbf{FID.} Fr\'echet Inception Distance measures the distance between the distributions
of real and generated motions~\cite{geng2025armflow}. It computes the Fr\'echet distance between the
activations of a pre-trained Inception network on real and generated samples. Lower FID
values indicate that the generated samples are closer to real samples in terms of their
distribution.
\item[$\bullet$]
\textbf{Diversity.} Diversity evaluates the variety of the generated human interactions~\cite{geng2025armflow}.
It measures how different the generated motions are from each other, ensuring that the
generative model produces a wide range of possible outcomes rather than repetitive or
similar ones. High diversity indicates a better performance in generating a rich set of
distinct motions.
\item[$\bullet$]
\textbf{Multimodality.} Multimodality evaluates whether the model can produce different types
of interaction motions for the same interactive entity, capturing the inherent variability in
human behavior~\cite{geng2025armflow}. High multimodality indicates that the model can produce diverse
outcomes across multiple distinct modes.
\item[$\bullet$]
\textbf{R Precision.} R Precision evaluates the proportion of relevant interaction motions
included in the top R generated results~\cite{guo2022generating}. It measures the accuracy of generation by
comparing the number of relevant motions within the first R results. This metric provides an
intuitive assessment of how well the model produces relevant outcomes in the top-ranked results.
\item[$\bullet$]
\textbf{MM Dist.} Multimodal Distance measures the similarity between each text prompt and its corresponding generated motion~\cite{liang2024intergen}. 
For each generated sample, it computes the Euclidean distance between the text embedding and the motion embedding generated from the same text, and reports the average distance over all samples. 
Lower values indicate better text-motion alignment.
\item[$\bullet$]
\textbf{Phase Transition Smoothness (PTS).} Phase Transition Smoothness measures the
smoothness of motion at phase boundaries in multi-phase interactions. It computes the
root-velocity jerk at each phase boundary as $\|\mathbf{v}_{t+1} - \mathbf{v}_{t-1}\|$,
where $\mathbf{v}_t$ denotes the root displacement at frame $t$, and averages over all
boundaries and both persons. Lower values indicate smoother transitions between consecutive
interaction phases.
\item[$\bullet$]
\textbf{Relative Distance Error (RDE).} Inter-Person Relative Distance Error
measures the accuracy of the spatial distance between the two generated persons. It computes
the mean absolute difference between the predicted root-to-root distance and the ground-truth
root-to-root distance across all frames.
\item[$\bullet$]
\textbf{Relative Orientation Error (ROE).} Inter-Person Relative Orientation Error
measures how accurately the generated person faces their interaction partner. For each frame, it
computes the cosine similarity between person~1's forward direction and the unit vector pointing
from person~1 toward person~2, and reports the mean absolute difference between the predicted and
ground-truth facing cosines across all frames. Lower values indicate that the generated persons
maintain more faithful relative orientations throughout the interaction.

\item[$\bullet$]
\textbf{Interpenetration Rate (IPR).} Interpenetration Rate measures the physical plausibility
of the generated interaction by quantifying body collisions between the two persons. Each body
joint is modeled as a sphere with an anatomically defined radius, and a frame is classified as
interpenetrating if any joint sphere of person~1 overlaps with any joint sphere of person~2.
The metric reports the fraction of frames exhibiting interpenetration. Lower values indicate
more physically plausible interactions.

\end{itemize}

\section{Modeling Derivation of Social-Structure-Centered HHI Generation}
\label{app:modeling_derivation}

\subsection{Planner-Executor Factorization}
Given a global interaction prompt \(y\), text-driven HHI generation aims to model the conditional distribution of a two-person motion sequence
\begin{equation}
p(\mathbf{X}^{1:T}\mid y), 
\qquad 
\mathbf{X}^{1:T}=\{\mathbf{x}^{(1)}_t,\mathbf{x}^{(2)}_t\}_{t=1}^{T}.
\end{equation}
We use the SMPL-based~\cite{SMPL_X_2019} motion representation, where the motion state of actor \(i\) at frame \(t\) is
\begin{equation}
\mathbf{x}^{(i)}_{t} =
\big(
\mathbf{r}^{(i)}_{t},
\mathbf{o}^{(i)}_{t},
\boldsymbol{\theta}^{(i)}_{t},
\boldsymbol{\eta}^{(i)}_{t}
\big),
\end{equation}
with \(\mathbf{r}^{(i)}_{t}\in\mathbb{R}^{3}\) the root translation, \(\mathbf{o}^{(i)}_{t}\in\mathbb{R}^{6}\) the root orientation, \(\boldsymbol{\theta}^{(i)}_{t}\in\mathbb{R}^{21\times 6}\) the body joint rotations, and \(\boldsymbol{\eta}^{(i)}_{t}\in\mathbb{R}^{22\times 3}\) the root-relative joint positions obtained by forward kinematics.

We introduce a latent social structure variable \(S\) as an intermediate representation that organizes the interaction before motion execution. 
Marginalizing over \(S\) gives
\begin{equation}
p(\mathbf{X}^{1:T}\mid y)
=
\sum_S p(\mathbf{X}^{1:T},S\mid y)
=
\sum_S p(\mathbf{X}^{1:T}\mid S,y)\,p(S\mid y).
\end{equation}
The term \(p(S\mid y)\) corresponds to social structure planning, while \(p(\mathbf{X}^{1:T}\mid S,y)\) corresponds to motion execution conditioned on the planned structure. 
We adopt the modeling approximation that, once \(S\) captures the phase progression and actor-specific role semantics, the global prompt provides largely redundant information for execution:
\begin{equation}
p(\mathbf{X}^{1:T}\mid S,y)
\approx
p_{\theta}(\mathbf{X}^{1:T}\mid S),
\qquad
p(S\mid y)\approx p_{\phi}(S\mid y).
\end{equation}
This yields the planner-executor formulation
\begin{equation}
p(\mathbf{X}^{1:T}\mid y)
\approx
\sum_S p_{\theta}(\mathbf{X}^{1:T}\mid S)\,p_{\phi}(S\mid y),
\end{equation}
where \(p_{\phi}\) is implemented by the LLM planner and \(p_{\theta}\) is implemented by the motion skill executor.

\subsection{Social Structure Derivation}
Since existing HHI datasets provide paired global text and two-person motion, we can derive \(S\) from paired motion data by constraining the LLM planner with phase-wise motion facts.

Given a paired HHI motion sequence \(\mathbf{X}^{1:T}\), we first detect the phase sequence
\begin{equation}
\mathcal{C}=f_{\mathrm{p}}(\mathbf{X}^{1:T}), 
\qquad 
\mathcal{C}=\{c_k\}_{k=1}^{K}, 
\qquad 
c_k\in\mathcal{V}_{\text{phase}},
\end{equation}
where
\begin{equation}
\mathcal{V}_{\text{phase}}
=
\{\texttt{approach},\texttt{contact},\texttt{release},\texttt{in\mbox{-}place}\}.
\end{equation}
The mapping \(f_{\mathrm{p}}(\cdot)\) is implemented by a forward state machine over motion-derived signals.

For each phase \(c_k\), we extract motion facts from the corresponding motion segment:
\begin{equation}
\mathcal{M}_k=\{m_k^{(1)},m_k^{(2)}\},
\qquad
m_k^{(i)}=
\big(
\delta_k,\,
u_k,\,
q_k,\,
\mathbf{a}_k^{(i)},\,
\mathbf{l}_k^{(i)}
\big),
\qquad i\in\{1,2\}.
\end{equation}
Here, \(\delta_k\) denotes inter-person distance evolution, \(u_k\) denotes the motion initiator, \(q_k\) captures contact semantics, \(\mathbf{a}_k^{(i)}\) summarizes the motion direction and facing state of actor \(i\), and \(\mathbf{l}_k^{(i)}\) describes limb-level cues such as reaching, arm lifting, bending, and foot activity.

Conditioned on the global interaction text \(y\), phase label \(c_k\), and motion facts \(\mathcal{M}_k\), the LLM planner produces actor-specific semantic descriptions:
\begin{equation}
\mathcal{Y}_k \sim p_{\phi}\!\left(\mathcal{Y}_k \mid y,\, c_k,\, \mathcal{M}_k\right),
\qquad
\mathcal{Y}_k=(y_k^{(1)},y_k^{(2)}).
\end{equation}
Finally, aggregating all phases yields the social structure used by the executor:
\begin{equation}
S=\{(c_k,y_k^{(1)},y_k^{(2)})\}_{k=1}^{K}.
\end{equation}
Although the final \(S\) contains only phase labels and actor-specific semantic descriptions, it is derived under motion fact constraints and is therefore aligned with the observed motion sequence.

\subsection{Phase-wise Motion Execution Modeling}
Given the planned social structure \(S\), the executor generates motion phase by phase:
\begin{equation}
p_{\theta}(\mathbf{X}^{1:T}\mid S)
\approx
\prod_{k=1}^{K}
\prod_{i=1}^{2}
p_{\theta}
\left(
\mathbf{x}_{k}^{(i)}
\mid
c_k,\,
y_k^{(i)},\,
\mathbf{p}_{k}^{(i)}
\right),
\end{equation}
where \(\mathbf{x}_{k}^{(i)}\) denotes actor \(i\)'s motion in phase \(k\), and \(\mathbf{p}_{k}^{(i)}\) is the partner-aware conditioning signal. 
Phase progression is grounded by self conditioning, where the anchor region of phase \(k\) is taken from the last \(e\) frames of the previous phase:
\begin{equation}
\mathbf{x}_{k}^{(i)}[1:e]
=
\mathbf{x}_{k-1}^{(i)}[T_{k-1}-e+1:T_{k-1}],
\end{equation}
where \(T_{k-1}\) denotes the duration of phase \(k-1\). 
Partner-aware coordination is grounded by representing the partner motion in the ego-centric frame of the current actor:
\begin{equation}
\mathbf{p}_{k}^{(i)}
=
R_{j\rightarrow i}\!\left(\mathbf{x}_{k}^{(j)}\right),
\qquad j\neq i.
\end{equation}

\subsection{Solo-to-Social Optimization}
We employ HY-Motion 1.0~\cite{hymotion2025} as the backbone of the motion executor and adapt it with LoRA. 
For selected linear layers with pretrained weight \(\mathbf{W}_0\), the adapted weight is
\begin{equation}
\mathbf{W}=\mathbf{W}_0+\Delta\mathbf{W},
\qquad
\Delta\mathbf{W}=\frac{\alpha}{r}\mathbf{B}\mathbf{A},
\end{equation}
where \(\mathbf{A}\) and \(\mathbf{B}\) are trainable low-rank matrices, \(r\) is the rank, and \(\alpha\) is the scaling factor.

The overall fine-tuning objective is
\begin{equation}
\mathcal{L}_{\mathrm{total}}
=
\mathcal{L}_{\mathrm{FM}}
+
w_{\mathrm{smooth}}\mathcal{L}_{\mathrm{smooth}}
+
w_{\mathrm{dist}}\mathcal{L}_{\mathrm{rel\_dist}}
+
w_{\mathrm{ori}}\mathcal{L}_{\mathrm{rel\_ori}}.
\end{equation}
Let \(\mathbf{x}_{\mathrm{gt}}\) denote the clean target motion, \(\boldsymbol{\epsilon}\sim\mathcal{N}(0,I)\) the Gaussian noise, and \(s\sim\mathcal{U}(0,1)\) the flow interpolation time. 
The noised motion is
\begin{equation}
\mathbf{x}_s=(1-s)\boldsymbol{\epsilon}+s\,\mathbf{x}_{\mathrm{gt}},
\end{equation}
and the flow-matching loss is
\begin{equation}
\mathcal{L}_{\mathrm{FM}}
=
\frac{1}{|\mathcal{V}|}
\sum_{t\in\mathcal{V}}
\left\|
v_{\theta}(\mathbf{x}_s,s)_t - (\mathbf{x}_{\mathrm{gt}}-\boldsymbol{\epsilon})_t
\right\|_2^2,
\end{equation}
where \(\mathcal{V}\) denotes the valid generated frames excluding anchor-prefix frames.

The smoothness loss is
\begin{equation}
\mathcal{L}_{\mathrm{smooth}}
=
\frac{1}{|\mathcal{V}|}
\sum_{t\in\mathcal{V}}
\left\|
\hat{\mathbf{x}}^{t} - \hat{\mathbf{x}}^{t-1}
\right\|_2^2.
\end{equation}
During training, the partner is treated as a fixed geometric reference using its ground-truth motion. 
The relative distance loss is
\begin{equation}
\mathcal{L}_{\mathrm{rel\_dist}}
=
\frac{1}{|\mathcal{V}|}
\sum_{t\in\mathcal{V}}
\left(
\left\|
\hat{\mathbf{r}}^{t}-\mathbf{r}_{p}^{t}
\right\|_2
-
\left\|
\mathbf{r}^{t}-\mathbf{r}_{p}^{t}
\right\|_2
\right)^2.
\end{equation}
The relative orientation loss is
\begin{equation}
\mathcal{L}_{\mathrm{rel\_ori}}
=
\frac{1}{|\mathcal{V}|}
\sum_{t\in\mathcal{V}}
\left(
\hat{\mathbf{f}}^{t} \cdot 
\mathbf{d}(\hat{\mathbf{r}}^{t},\mathbf{r}_{p}^{t})
-
\mathbf{f}^{t} \cdot
\mathbf{d}(\mathbf{r}^{t},\mathbf{r}_{p}^{t})
\right)^2,
\end{equation}
where
\begin{equation}
\mathbf{d}(a,b)=\frac{b-a}{\|b-a\|_2+\varepsilon}
\end{equation}
is the normalized direction vector from \(a\) to \(b\). 
Together, these objectives adapt the solo motion backbone into a socially aware executor by preserving atomic motion quality, improving temporal smoothness, and enforcing partner-relative coordination.

\begin{figure}[t]
    \centering
    \includegraphics[width=1\linewidth]{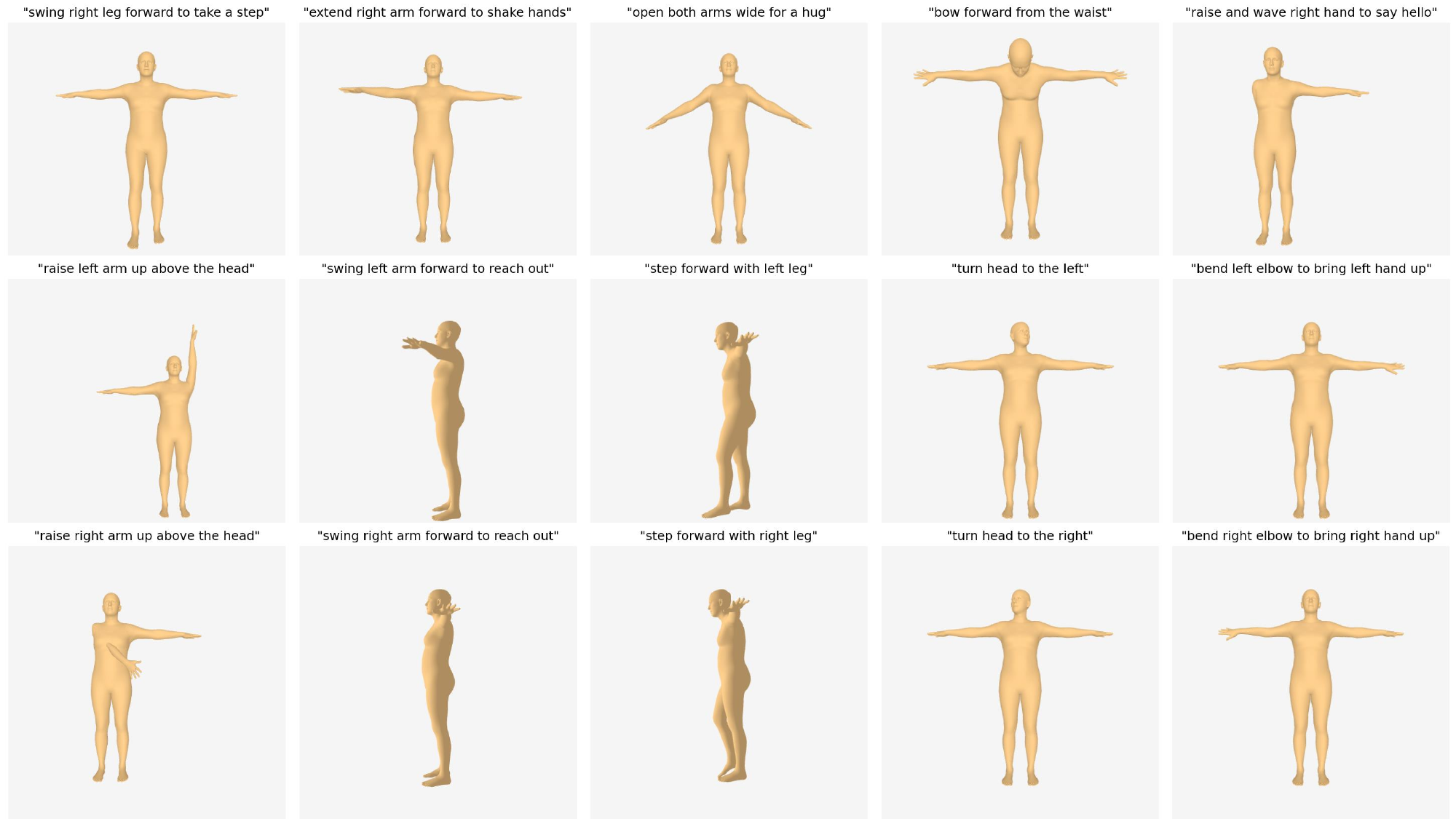}
    \caption{Visualizations of LLM-based human atomic motion execution. Given atomic action prompts, the LLM can sometimes recover coarse pose-level semantics, but the generated motions are often static, low-amplitude, and lack continuous dynamics. This supports our observation that LLMs can reason about social structure planning but are insufficient for direct motion execution.}
    \label{fig:llm_atomic_motion_execution}
\end{figure}

\section{LLM-based Human Atomic Motion Execution}
\label{app:llm_atomic_motion_execution}
To examine whether LLMs can directly serve as motion executors, we ask the LLM to control SMPL-based human motions from a set of atomic action prompts, such as stepping forward, raising an arm, reaching out, bowing, and waving. 
These prompts describe simple single-person motions and therefore provide a basic test of motion execution ability before considering more complex two-person interactions.

As shown in Fig.~\ref{fig:llm_atomic_motion_execution}, the LLM can sometimes recover coarse pose-level semantics, such as lifting the specified arm or orienting the body toward the intended direction. 
However, the generated results are often close to static poses, with limited motion amplitude, weak temporal evolution, and insufficient coordination among body joints. 
For locomotion-related prompts, the model frequently fails to produce dynamic displacement or realistic stepping motion. 
These observations indicate that LLMs may understand the semantic meaning of atomic motion descriptions, but they do not reliably generate continuous, dynamic, and physically plausible 3D motion. 
This further supports our planner-executor design: LLMs are suitable for semantic-level planning, while motion execution should be handled by specialized motion models.

\section{Ablation Study Details}
\label{app:ablation_visualization}
We provide additional details on the ablation settings and qualitative comparisons. 
The ablations are designed to examine whether the proposed improvements come from social structure modeling and grounding, rather than from using an LLM alone. 
Specifically, \textit{w/o social structure planning} removes the phase-wise social structure and uses only the original global text, testing whether explicit phase and role supervision is necessary. 
\textit{w/o motion facts} keeps LLM planning but removes motion-derived constraints, testing whether language-only plans can remain aligned with the actual motion. 
\textit{w/o self conditioning} removes the previous-phase anchor prefix, testing whether phase progression can be grounded into temporally continuous motion. 
\textit{w/o partner conditioning} removes ego-relative partner motion input, testing whether partner-aware coordination can be grounded into inter-person geometry.

\begin{figure}
    \centering
    \includegraphics[width=1\linewidth]{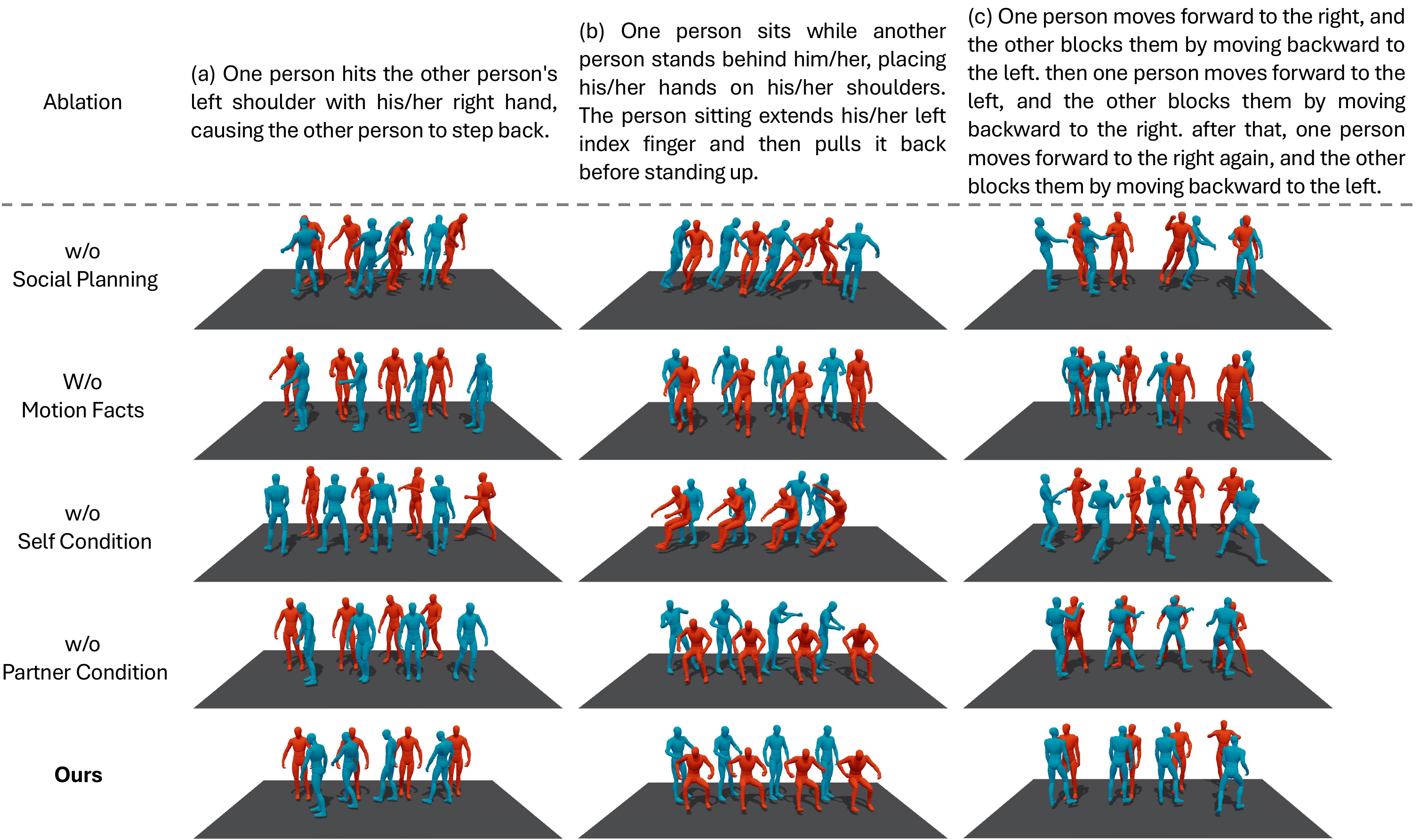}
     \caption{Qualitative ablation results. 
    Removing social structure planning or motion facts weakens semantic faithfulness and motion-aligned phase structure. 
    Removing self conditioning degrades phase continuity, while removing partner conditioning harms inter-person coordination. 
    Our full model produces more coherent phase progression and stronger partner-aware coordination.}
    \label{fig:ablation_vis}
\end{figure}

As shown in Fig.~\ref{fig:ablation_vis}, without social structure planning, the generated motions often show weak semantic faithfulness to the interaction prompt. 
Without motion facts, the phase decomposition and role assignment may appear plausible in language but become less aligned with the actual motion dynamics. 
For motion grounding, removing self conditioning leads to less stable phase transitions, while removing partner conditioning weakens relative distance, orientation, and response coordination between the two actors. 
In contrast, our full model better preserves phase progression and partner-aware coordination across the shown interaction scenarios.

\section{User Study}
\label{app:user_study}

\begin{table}[t]
\centering
\caption{User study questionnaire. 
We evaluate generated HHI motions along four dimensions: phase decomposition accuracy, global text-motion alignment, partner coordination, and phase-level alignment. 
Participants judge whether the phase-level prompts faithfully reflect the global interaction description and whether the generated motions correctly realize the intended interaction, temporal phase progression, and partner-aware coordination.}
\label{tab:user_study_questions}
\small
\setlength{\tabcolsep}{4pt}
\renewcommand{\arraystretch}{1.2}
\begin{tabularx}{\linewidth}{p{0.30\linewidth}X}
\toprule
\textbf{Dimension} & \textbf{Question} \\
\midrule
(a) Phase Decomposition Accuracy &
How accurately do the phase-level text prompts decompose the global text description? \\
\midrule
(b) Global Text-Motion Alignment &
How well does the generated full motion sequence align with the global text description? Consider whether the interaction, movement direction, action intention, and final outcome correspond to the global prompt. \\
\midrule
(c) Partner Coordination &
How well do the two partners coordinate during the interaction? Consider whether they respond appropriately, maintain reasonable spatial relationships, and perform coordinated movements. \\
\midrule
(d) Phase-Level Alignment &
How well does each generated motion phase align with its corresponding phase-level text description? Consider the phase type, i.e., approach, contact, release, or in-place, and whether the intended phase-specific actions occur at the appropriate time. \\
\bottomrule
\end{tabularx}
\end{table}

\begin{figure}[t]
    \centering
    \includegraphics[width=1\linewidth]{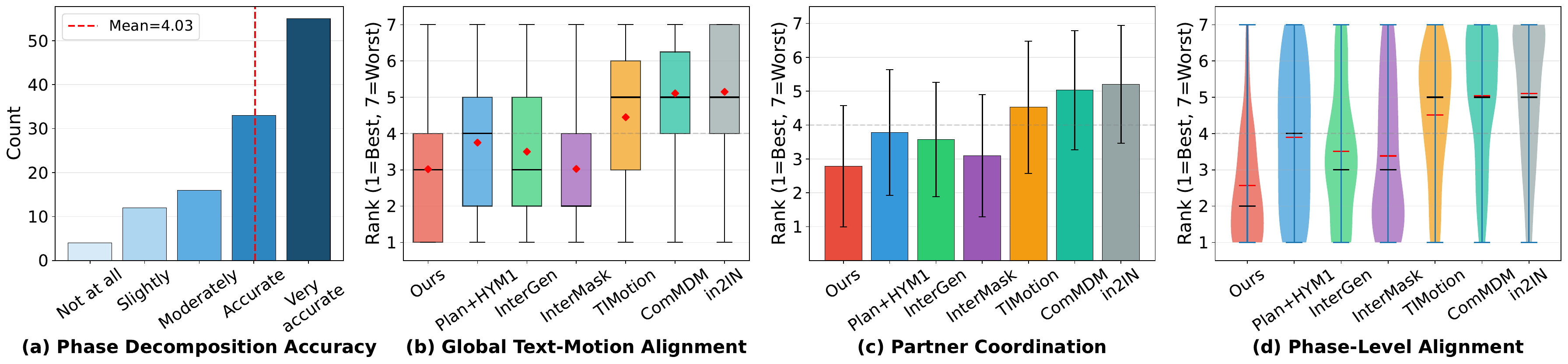}
    \caption{User study results across four evaluation aspects. 
    (a) Phase decomposition accuracy of the LLM-generated phase-level prompts. 
    (b) Global text-motion alignment between the generated full motion sequence and the global prompt. 
    (c) Partner coordination between the two actors. 
    (d) Phase-level alignment between generated motions and decomposed phase prompts.}
    \label{fig:questionnaire_result}
\end{figure}

\begin{figure}[H]
    \centering
    \includegraphics[width=0.95\linewidth]{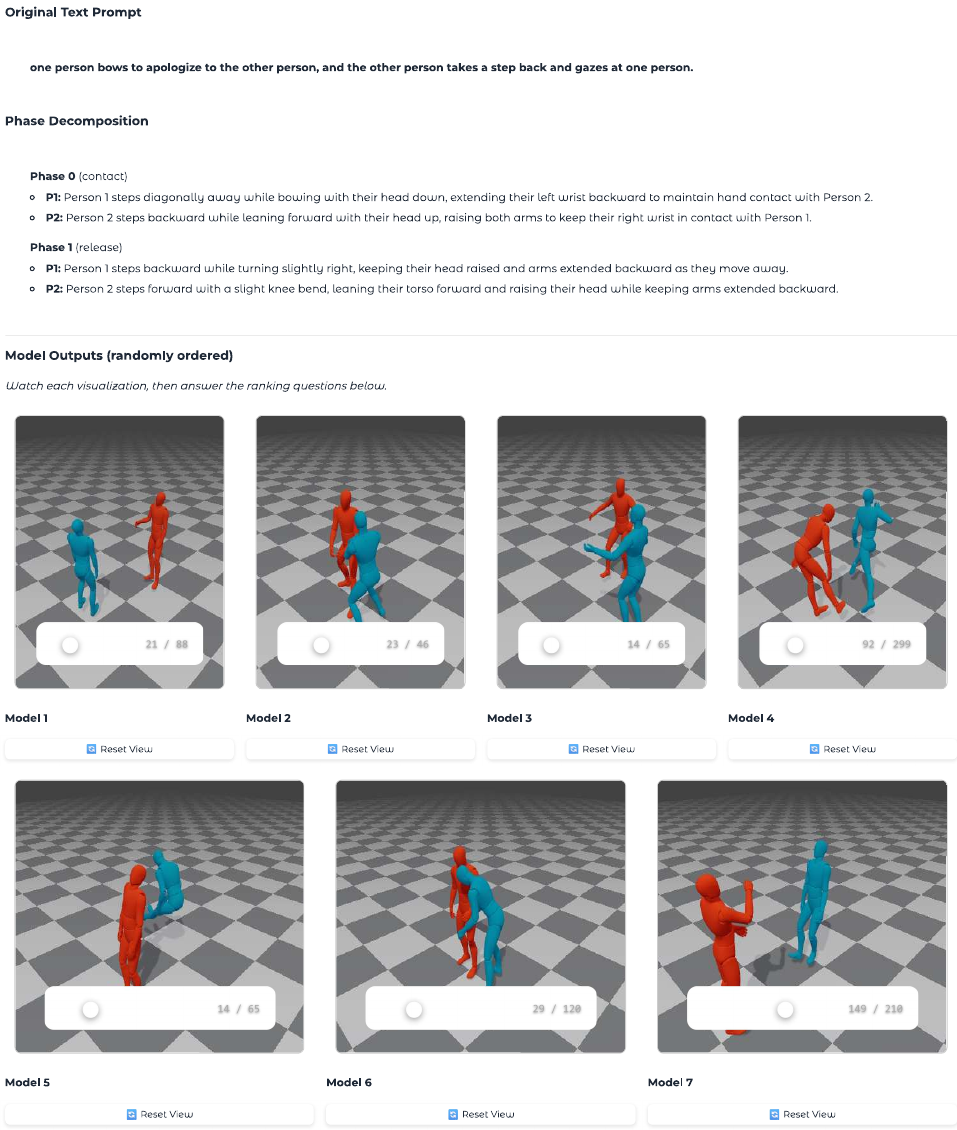}
    \caption{User study interface. 
    Each case presents the original global text prompt, the LLM-decomposed phase-level prompts, and seven randomly ordered model outputs. 
    Participants first evaluate the accuracy of the phase-level prompts, and then rank the generated motions according to global text-motion alignment, partner coordination, and phase-level alignment. 
    The model order is randomized for each case to reduce ordering bias.}
    \label{fig:questionnaire}
\end{figure}

We conducted a user study with a Gradio-based interface, as illustrated in Fig.~\ref{fig:questionnaire}. 
Each participant evaluated 10 randomly selected cases, resulting in 120 case evaluations from 12 participants in total. 
All participants had research backgrounds in computer vision or closely related areas. 
For each case, participants viewed animations generated by seven models and answered the questions listed in Table~\ref{tab:user_study_questions}. 
Participants first rated the phase decomposition accuracy of the phase-level prompts on a 5-point Likert scale, where 1 denotes ``not accurate at all'' and 5 denotes ``very accurate.'' 
They then evaluated the generated animations along three motion-related aspects: global text-motion alignment, partner coordination, and phase-level alignment. 
For these three aspects, participants ranked the seven methods from 1 (best) to 7 (worst). 
To reduce ordering bias, the presentation order of the seven animations was randomly shuffled for each case.

The results are summarized in Fig.~\ref{fig:questionnaire_result}. 
Fig.~\ref{fig:questionnaire_result}(a) shows that most responses for phase decomposition accuracy fall into the ``Accurate'' and ``Very accurate'' categories, with a mean score of 4.03, indicating that the LLM-generated phase prompts are generally faithful to the original global descriptions. 
Fig.~\ref{fig:questionnaire_result}(b) reports the ranking distribution for global text-motion alignment, where our method achieves the most favorable overall rankings, suggesting stronger faithfulness to the intended interaction semantics. 
Fig.~\ref{fig:questionnaire_result}(c) shows that our method obtains the best average ranking for partner coordination, indicating stronger mutual responsiveness and more coherent inter-person spatial relationships. 
Fig.~\ref{fig:questionnaire_result}(d) further shows that our method better follows the decomposed phase-level prompts, preserving clearer temporal structure across interaction phases. 
Overall, the user study supports our quantitative findings and demonstrates improvements in global text-motion faithfulness, partner-aware coordination, and phase-level temporal consistency.

\section{Social Structure Planning Results}
\label{app:social_planning}

We present the prompt demonstration used for LLM-based social structure planning in Fig.~\ref{fig:social_structure_prompt}. 
The prompt asks the LLM to convert a global two-person interaction description into a phase-wise plan, where each phase is assigned one of four interaction types, \texttt{approach}, \texttt{contact}, \texttt{release}, or \texttt{in-place}, together with partner-aware action descriptions for P1 and P2. 
This format makes the implicit temporal progression and role coordination in the global prompt explicit, enabling the planned social structure to serve as structured supervision for interaction motion generation.

Table~\ref{tab:phase-level prompts} provides the social structure planning results for the eight qualitative cases shown in Fig.~\ref{fig:main_quali_res}. 
For each global interaction prompt, the LLM planner decomposes the interaction into one or more temporally ordered phases and assigns partner-aware action descriptions to the two actors within each phase. 
These phase-level plans make the implicit interaction organization explicit, specifying both how the interaction progresses over time and how the two actors should coordinate their roles. 
They are used as structured conditions for interaction motion execution. 

Table~\ref{tab:llm_planning_comparison} provides planning results across different LLMs under the same global interaction prompt. 
Different LLMs recover broadly consistent interaction structure, including approach from behind, shoulder contact, and response from the seated actor, but they vary in phase completeness and action granularity. 
This supports the usefulness of LLMs for semantic social structure planning, while also motivating the use of motion-derived facts to constrain planning and improve motion alignment.

\section{Discussion}
\label{app:SocialImpacts}

\mypara{Future Work}
This work formulates text-driven 3D human-human interaction generation as social structure modeling and grounding, but several directions remain open. 
First, our current framework focuses on two-person interactions. 
Extending social structure modeling to multi-person scenarios is an important future direction, where interaction organization may involve group roles, changing subgroups, and more complex coordination patterns. 
Second, our phase vocabulary captures common interaction stages such as approach, contact, release, and in-place coordination, but real human interactions can involve more subtle temporal and social dynamics. 
Future work may explore richer or adaptive social structure representations, uncertainty-aware planning, and more diverse interaction data that better cover different body types, cultural contexts, and social behaviors. 

\mypara{Societal Impact}
This work may have positive impact by enabling more controllable and semantically grounded generation of two-person human interactions. 
It can benefit animation production, virtual avatars, embodied AI, social robotics, human-computer interaction, and simulation environments for training or education by reducing manual motion authoring effort and supporting richer virtual social scenarios. 
Potential negative impacts should also be considered. 
Generated human interactions may be misused to create misleading synthetic content, and models trained on existing motion datasets may inherit biases in body types, interaction styles, cultural norms, or social behaviors. 
Moreover, representing interaction through planned phases and roles may simplify the complexity of real human social behavior, especially in ambiguous, emotional, or culturally specific settings. 
The method does not involve private personal data or high-risk deployment, but responsible use and careful evaluation remain necessary when applying generated interaction motions in social or human-facing contexts.


\begin{figure}[t]
\centering
\begin{minipage}{0.95\linewidth}
\begin{tcblisting}{
  colback=gray!5,
  colframe=black!50,
  title=LLM Prompt for Social Structure Planning,
  fonttitle=\bfseries,
  listing only,
  enhanced,
  listing options={
    basicstyle=\ttfamily\scriptsize,
    breaklines=true,
    columns=fullflexible,
    keepspaces=true,
    showstringspaces=false
  }
}
You are a professional social psychology expert specializing in analyzing the social behaviors of human-human interactions.

Your task is to convert a global two-person interaction description into a phase-wise social structure plan.
The plan should describe how the interaction unfolds over time and what each person does in each phase.

Definitions:
- Person 1 / P1 refers to the first person mentioned in the global prompt.
- Person 2 / P2 refers to the other person.
- Use only the following phase types:
  1. approach: one or both people move closer or prepare to interact before contact.
  2. contact: physical contact, object transfer, blocking, supporting, or direct interaction occurs.
  3. release: physical contact or interaction ends and one or both people withdraw or return to neutral.
  4. in-place: the interaction happens mostly without clear approach/contact/release progression,
     or the actors coordinate while staying near their positions.

Instructions:
1. Decompose the global prompt into 1 to 4 temporally ordered phases.
2. Assign exactly one phase type to each phase: approach, contact, release, or in-place.
3. For each phase, write one action sentence for P1 and one action sentence for P2.
4. The P1 and P2 actions must be partner-aware.
5. Preserve the semantics of the global prompt. Do not invent unrelated actions.
6. If the prompt implies asymmetric roles, make them explicit, such as initiator/receiver, attacker/defender, giver/taker, or supporter/assisted person.
7. If there is no clear movement toward or away from the partner, use in-place.
8. Keep each action concise, physically plausible, and suitable for motion generation.

Output format:
Phase 1: (<phase_type>)
P1 action: <one sentence describing Person 1's action>
P2 action: <one sentence describing Person 2's action>
Phase 2: (<phase_type>)
P1 action: <one sentence describing Person 1's action>
P2 action: <one sentence describing Person 2's action>

Global interaction prompt:
"<INPUT_GLOBAL_PROMPT>"
\end{tcblisting}
\end{minipage}
\caption{Prompt used for LLM-based social structure planning.}
\label{fig:social_structure_prompt}
\end{figure}

\begin{table}[t]
\centering
\caption{Social structure prompts used for the qualitative examples in Fig.~\ref{fig:main_quali_res}. 
All methods are conditioned on the global interaction prompt shown in the second column. 
Only Our Planning + HYM1 and Ours use the phase-wise social structure text shown in the third column. }
\label{tab:phase-level prompts}
\small
\setlength{\tabcolsep}{4pt}
\renewcommand{\arraystretch}{1.15}
\resizebox{0.94\textwidth}{!}{
\begin{tabularx}{\linewidth}{p{0.06\linewidth}p{0.28\linewidth}X}
\toprule
\textbf{ID} & \textbf{Global Interaction Text} & \textbf{Social Structure Planning Result} \\
\midrule

a &
The first one's right hand pushes the other person's left shoulder, and the other person blocks it with their hand. &
\textbf{Phase 1: (contact)}\newline
\textbf{P1 action:} Person 1 pushes Person 2's left shoulder with their right hand while stepping back.\newline
\textbf{P2 action:} Person 2 blocks the push with their hand and steps back to create distance. \\
\midrule

b &
The two are photographing themselves together. &
\textbf{Phase 1: (in-place)}\newline
\textbf{P1 action:} Person 1 holds or positions the camera/phone in front of both people and poses for the photo.\newline
\textbf{P2 action:} Person 2 stays close to Person 1 and poses together for the photo. \\
\midrule

c &
One retrieves a paper from one's pocket and passes it to the other. &
\textbf{Phase 1: (approach)}\newline
\textbf{P1 action:} Person 1 walks forward and extends an arm to retrieve a paper from their pocket.\newline
\textbf{P2 action:} Person 2 stands still facing Person 1, waiting for the paper to be passed.\newline
\textbf{Phase 2: (contact)}\newline
\textbf{P1 action:} Person 1 passes the paper into Person 2's hand.\newline
\textbf{P2 action:} Person 2 grasps and receives the paper from Person 1. \\
\midrule

d &
The first person raises his/her hands, reaches forward, and pinches both cheeks of the second person. &
\textbf{Phase 1: (approach)}\newline
\textbf{P1 action:} Person 1 raises both hands and extends both arms forward toward Person 2's face.\newline
\textbf{P2 action:} Person 2 remains mostly still, facing Person 1 and waiting for the interaction.\newline
\textbf{Phase 2: (contact)}\newline
\textbf{P1 action:} Person 1 places both hands on Person 2's cheeks and gently pinches both sides of the face.\newline
\textbf{P2 action:} Person 2 stays in place while receiving the cheek-pinching interaction, with only minor head or upper-body response.\newline
\textbf{Phase 3: (release)}\newline
\textbf{P1 action:} Person 1 releases Person 2's cheeks and slightly withdraws both hands from Person 2's face.\newline
\textbf{P2 action:} Person 2 remains facing Person 1 and returns to a neutral posture after the contact ends. \\
\midrule

e &
The first person advances and launches an attack with the right hand while the other person retreats. &
\textbf{Phase 1: (in-place)}\newline
\textbf{P1 action:} Person 1 advances toward Person 2 with a raised right hand, preparing to strike.\newline
\textbf{P2 action:} Person 2 retreats while facing Person 1, maintaining a defensive posture without contact. \\
\midrule

f &
Two people reach out their right hands to greet one another. &
\textbf{Phase 1: (contact)}\newline
\textbf{P1 action:} Person 1 walks forward extending their right hand to initiate a greeting.\newline
\textbf{P2 action:} Person 2 moves slightly forward extending their right hand to meet Person 1. \\
\midrule

g &
The other person assists one person by lifting them up. &
\textbf{Phase 1: (contact)}\newline
\textbf{P1 action:} Person 1 holds or supports Person 2 and lifts them upward.\newline
\textbf{P2 action:} Person 2 is physically supported by Person 1 and is lifted upward with assistance.\newline
\textbf{Phase 2: (release)}\newline
\textbf{P1 action:} Person 1 gradually reduces support and releases Person 2 after the lift.\newline
\textbf{P2 action:} Person 2 returns to a stable standing posture after being lifted. \\
\midrule

h &
One of the persons suddenly stands up while both were sitting. &
\textbf{Phase 1: (in-place)}\newline
\textbf{P1 action:} Person 1 suddenly rises from a sitting position to a standing posture.\newline
\textbf{P2 action:} Person 2 remains seated and stays mostly still while Person 1 stands up. \\

\bottomrule
\end{tabularx}
}
\end{table}

\begin{table}[t]
\centering
\small
\caption{Comparison of social structure planning results produced by different LLMs for the same global interaction text. 
All models are prompted to decompose the interaction into phase-level structures and assign partner-aware actions to P1 and P2.}
\label{tab:llm_planning_comparison}
\begin{tabularx}{\linewidth}{p{0.16\linewidth}X}
\toprule
\textbf{LLM Model} & \textbf{Social Structure Planning Result} \\
\midrule
\multicolumn{2}{p{\linewidth}}{
\textbf{Global interaction text:} One person sits while the other person bends from behind, patting the person's shoulder with both hands, and the person raises his/her right hand.
} \\
\midrule

ChatGPT 5.4 &
\textbf{Phase 1: (approach)}\newline
\textbf{P1 action:} P1 sits in place while slightly orienting the upper body as P2 approaches from behind.\newline
\textbf{P2 action:} P2 moves behind P1 and bends forward toward P1's shoulders.\newline
\textbf{Phase 2: (contact)}\newline
\textbf{P1 action:} P1 remains seated and raises the right hand in response to P2's shoulder pat.\newline
\textbf{P2 action:} P2 bends from behind P1 and pats P1's shoulder with both hands.\newline
\textbf{Phase 3: (release)}\newline
\textbf{P1 action:} P1 keeps seated and lowers the right hand back toward a neutral position.\newline
\textbf{P2 action:} P2 stops patting P1's shoulder and begins to withdraw the hands.
\\
\midrule

Gemini 3 &
\textbf{Phase 1: (approach)}\newline
\textbf{P1 action:} Person 1 sits in a neutral, stationary posture as the receiver, awaiting the interaction from behind.\newline
\textbf{P2 action:} Person 2 acts as the initiator by bending their torso forward from directly behind Person 1, preparing to reach out with both arms.\newline
\textbf{Phase 2: (contact)}\newline
\textbf{P1 action:} Person 1 remains seated and raises their right hand upward toward their shoulder level in a responsive gesture to the touch.\newline
\textbf{P2 action:} Person 2 makes physical contact by placing both hands on Person 1's shoulders and executing a gentle, repeated patting motion.
\\
\midrule

Qwen3.5 &
\textbf{Phase 1: (approach)}\newline
\textbf{P1 action:} P1 sits still in a chair, facing forward while P2 approaches.\newline
\textbf{P2 action:} P2 moves behind P1 and bends their torso down towards P1's shoulders.\newline
\textbf{Phase 2: (contact)}\newline
\textbf{P1 action:} P1 raises their right hand while remaining seated.\newline
\textbf{P2 action:} P2 places both hands on P1's shoulders and pats them gently.\newline
\textbf{Phase 3: (release)}\newline
\textbf{P1 action:} P1 lowers their right hand back to a neutral resting position.\newline
\textbf{P2 action:} P2 stops patting and straightens their body to withdraw hands from P1.
\\
\midrule

DeepSeek-V4 &
\textbf{Phase 1: (approach)}\newline
\textbf{P1 action:} P1 remains still, facing forward and unaware of the approaching touch.\newline
\textbf{P2 action:} P2 walks up behind the seated person, leaning forward with both hands extended toward the seated person's shoulders.\newline
\textbf{Phase 2: (contact)}\newline
\textbf{P1 action:} P1 responds to the shoulder pats by raising their right hand upward in acknowledgment or greeting.\newline
\textbf{P2 action:} P2 places both hands on the seated person's shoulders and pats them gently in a friendly gesture.\newline
\textbf{Phase 3: (release)}\newline
\textbf{P1 action:} P1 lowers their right hand back down to a resting position.\newline
\textbf{P2 action:} P2 lifts both hands off the seated person's shoulders and straightens back up.
\\

\bottomrule
\end{tabularx}
\end{table}

\end{document}